\documentclass[11pt]{article}

\usepackage[preprint]{acl}

\usepackage{times}
\usepackage{latexsym}

\usepackage[T1]{fontenc}
\usepackage[utf8]{inputenc}

\usepackage{microtype}

\usepackage{inconsolata}

\usepackage{graphicx}

\usepackage{enumitem}
\usepackage[table,xcdraw]{xcolor}
\usepackage{booktabs}

\usepackage{amsmath}

\usepackage{multirow}
\usepackage{tcolorbox}
\usepackage{listings}
\usepackage{fontawesome}
\usepackage{hyperref}

\lstdefinestyle{PromptStyle}{
    basicstyle=\scriptsize\ttfamily\color{black!85},
    frame=lr,
    rulecolor=\color{black!60},
    backgroundcolor=\color{black!2},
    framesep=4pt,
    breaklines=true,
    breakatwhitespace=false,
    columns=fullflexible,
    numbers=none,
    aboveskip=6pt,
    belowskip=6pt,
    xleftmargin=2pt,
    xrightmargin=2pt,
    upquote=true
}

\newcommand{\ie}{\emph{i.e., }}
\newcommand{\eg}{\emph{e.g., }}

\definecolor{darkgreen}{HTML}{1B5E20} % 深绿色（非常稳重）
\title{PERM: Psychology-grounded Empathetic Reward Modeling for Large Language Models}
\author{
 \textbf{Chengbing Wang\textsuperscript{1}\thanks{Equal contribution-\{wcb0219,qqqqqzheng\}@gmail.com}},
 \textbf{Wuqiang Zheng\textsuperscript{1}\footnotemark[1]},
 \textbf{Yang Zhang\textsuperscript{2}}, 
 \textbf{Fengbin Zhu\textsuperscript{2}},
\\
 \textbf{Junyi Cheng\textsuperscript{3}},
 \textbf{Yi Xie\textsuperscript{3}},
 \textbf{Wenjie Wang\textsuperscript{1}\thanks{Corresponding author-wenjiewang96@gmail.com}},
 \textbf{Fuli Feng\textsuperscript{1}},
\\
\\
 \textsuperscript{1}University of Science and Technology of China, \\
 \textsuperscript{2}National University of Singapore, 
 \textsuperscript{3}Huawei Technologies
\\
}

\begin{document}
\maketitle
\begin{abstract}

Large Language Models (LLMs) are increasingly deployed in human-centric applications, yet they often fail to provide substantive emotional support. While Reinforcement Learning (RL) has been utilized to enhance empathy of LLMs, existing reward models typically evaluate empathy from a single perspective, overlooking the inherently bidirectional interaction nature of empathy between the supporter and seeker as defined by \textit{Empathy Cycle} theory. To address this limitation, we propose \textit{\textbf{P}sychology-grounded \textbf{E}mpathetic \textbf{R}eward \textbf{M}odeling} (PERM). PERM operationalizes empathy evaluation through a bidirectional decomposition: 1) Supporter perspective, assessing internal resonation and communicative expression; 2) Seeker perspective, evaluating emotional reception. Additionally, it incorporates a bystander perspective to monitor overall interaction quality. Extensive experiments on a widely-used emotional intelligence benchmark and an industrial daily conversation dataset demonstrate that PERM outperforms state-of-the-art baselines by over 10\%. Furthermore, a blinded user study reveals a 70\% preference for our approach, highlighting its efficacy in generating more empathetic responses\footnote{\faGithub\ \href{https://github.com/ZhengWwwq/PERM}{https://github.com/ZhengWwwq/PERM}.}.

\end{abstract}

\section{Introduction}

% \begin{figure}[!ht]
%     \centering
%     \includegraphics[width=0.99\linewidth]{figures/intro_fig.pdf}
%     \caption{An example illustrating the effectiveness of different reward modeling methods. Modeling only the LLM's responses leads to generic, surface-level expressions, while modeling only the user's needs yields solution-oriented but emotionally cold responses. By jointly considering both perspectives, PERM enables more authentic and comprehensive empathetic replies.}
%     \label{fig:intro}
% \end{figure}

As Large Language Models (LLMs) are increasingly deployed in human-centric applications~\cite{LLM-survey,LLM-survey2}, their role in supporting psychological well-being has become a focal point of concern~\cite{intro-1-Top100GenAI,intro-1-impcase0,intro-1-impcase1,intro-1-impcase2,intro-1-impcase3}. When users disclose psychological vulnerabilities or sensitive personal beliefs, there is a functional expectation for these LLMs to generate empathetic responses that provide substantive emotional support. However, prior studies have shown that state-of-the-art LLMs often fail to meet these expectations, exhibiting limited empathetic understanding and inappropriate response strategies~\cite{intro-1-badcase1,intro-1-badcase2,intro-1-badcase3,intro-1-badcase4}. 
% These limitations underscore the need for systematic methodological advances to enhance the empathetic capabilities of LLMs.
These limitations highlight the need to enhance LLMs’ empathetic capabilities. 

\begin{figure}[t]
    \centering
    \includegraphics[width=0.99\linewidth]{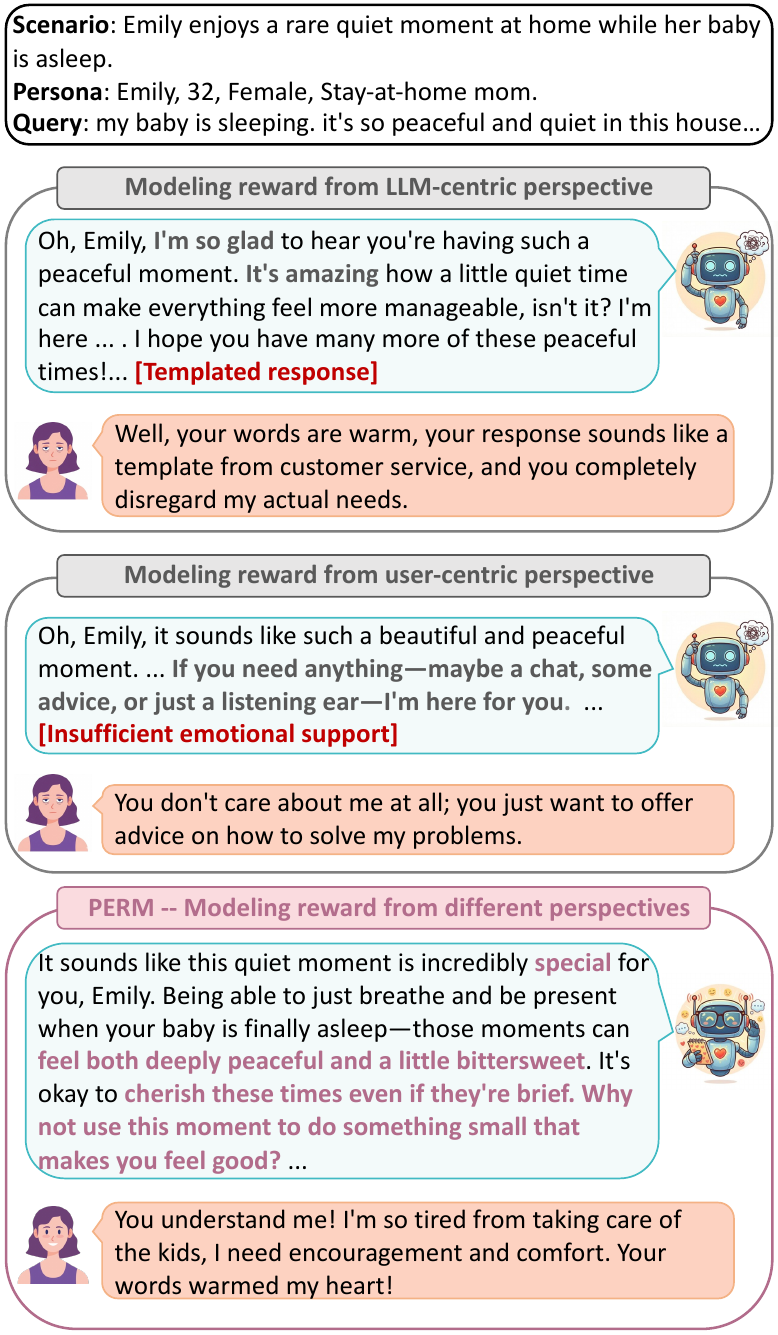}
    \caption{An example illustrating the effectiveness of different reward modeling methods. Modeling only the LLM's responses leads to generic, surface-level expressions, while modeling only the user's needs yields solution-oriented but emotionally cold responses. By jointly considering both perspectives, PERM enables more authentic and comprehensive empathetic replies.}
    \label{fig:intro}
\end{figure}

% ~\cite{sharma2021towards, wang2025rlver}

Early research on enhancing LLM empathy primarily utilized Supervised Fine-Tuning (SFT)~\cite{sun2021psyqa, liu2023chatcounselor}. However, SFT is constrained by its reliance on high-quality demonstrations and often generalizes poorly to diverse inputs~\cite{chu2025sft}. Later, research transitioned toward Reinforcement Learning (RL) to address these bottlenecks. Unlike the token-matching objective of SFT, RL facilitates improvement through exploration, optimizing reward functions that operationalize empathy via user-centric signals (\eg feedback)~\cite{wang2025rlver} or response-centric attributes (\eg warmth)~\cite{sharma2021towards}. By optimizing qualitative objectives rather than relying on token-matching, RL–based methods demonstrate superior generalization performance in complex psychological support scenarios.

Despite the promise, we argue that current RL-based approaches share a fundamental limitation: their reward modeling fails to model empathy through a bidirectional interaction. As established in \textit{Empathy Cycle} theory~\cite{barrett1981empathy} from \textit{psychology}, empathy is not a standalone output but a closed loop between the supporter (the LLM) and the seeker (the user). The supporter resonates with and expresses emotions to the seeker, and the cycle is only complete when the seeker perceives and reacts to that expression.  
This bidirectional nature suggests that empathetic reward modeling should jointly account for both the supporter’s reasonation/expression and the seeker’s reception. 
However, current approaches typically evaluate empathy from a single perspective—focusing on either the supporter's response or the seeker's feedback in isolation. 
By ignoring the mutual alignment, these methods fail to close the \textit{Empathy Cycle}, often resulting in templated responses or insufficient emotional support, as shown in Figure~\ref{fig:intro}.

To address this limitation, we reformulate empathetic reward modeling through the lens of the \textit{Empathy Cycle} and introduce \textit{\textbf{P}sychology-grounded \textbf{E}mpathic \textbf{R}eward \textbf{M}odeling} (PERM) framework. This framework explicitly decomposes empathetic rewards into a bidirectional evaluation of both the seeker (user) and the supporter (LLM):

\begin{itemize}[leftmargin=*]
    \item \textbf{Supporter's perspective. }PERM evaluates two complementary dimensions: \textit{Resonation}, capturing the supporter's understanding of the seeker’s emotions and implicit needs, and \textit{Expression}, assessing how effectively this understanding is communicated through appropriate empathetic qualities, such as warmth.
    \item \textbf{Seeker's perspective. }PERM evaluates \textit{Reception}, assessing whether the supporter's response is perceived as supportive and well aligned with the seeker’s needs.

\end{itemize}
To further ensure interaction quality, PERM additionally incorporates a \textbf{bystander perspective}. This perspective monitors non-empathetic aspects of the interaction, such as coherence and relevance, providing an additional signal to align rewards with realistic human needs.

To evaluate the effectiveness of PERM, we employ it to fine-tune LLMs with RL on the EmpatheticDialogues dataset~\cite{rashkin2019towards} and assess their performance across three settings: a general emotional intelligence benchmark, a daily conversations benchmark, and a real-world user study.
Across all evaluations, PERM consistently outperforms baselines, yielding over 10\% improvements on emotional Intelligence and conversation benchmarks and being preferred by over 70\% of participants in the user study, demonstrating substantial gains in both empathetic quality and overall emotional intelligence.

Our key contributions are as follows:

\begin{itemize}[leftmargin=*]
    \item We identify the limitations of single-perspective reward modeling and advocate a shift toward psychologically grounded, bidirectional empathy modeling for reward learning, inspired by \textit{Empathy Cycle} theory.
    \item We propose PERM, a reward modeling framework that evaluates empathy from multiple complementary perspectives—including supporter, seeker, and bystander—yielding a more comprehensive assessment of empathetic behavior.
    \item We demonstrate the effectiveness of PERM through consistent improvements in empathy and overall emotional intelligence across an emotional intelligence benchmark, a daily conversation benchmark, and a user study.
\end{itemize}

% \section{Task Formulation}

% We study empathetic response generation with LLMs. Let  
% \(\mathcal{D} = \{x_i\}_{i=1}^{N}\) denote a set of empathy-seeking queries. Given an input \(x\), an LLM generates a response \(\hat{y}\), which is evaluated by a reward model (RM):
% \begin{equation}
% x \xrightarrow{\text{LLM}} \hat{y} \xrightarrow{\text{RM}} r,
% \end{equation}
% where \(r \in \mathbb{R}\) measures the empathetic quality of \(\hat{y}\).

% The training objective is to learn a policy \(\pi_\theta\) that maximizes the expected reward:
% \begin{equation}
% \max_{\theta} \;
% \mathbb{E}_{x \sim \mathcal{D}, \, \hat{y} \sim \pi_\theta(\cdot \mid x)}
% \left[ \text{RM}(x, \hat{y}) \right].
% \end{equation}

% However, when the reward model provides an incomplete characterization of real-world empathy, optimizing this objective may not lead to genuine empathy.

% This misalignment highlights a key limitation of reward modeling for empathetic response generation and motivates the need for more complete reward formulations.

\section{PERM Framework}

\begin{figure*}[t]
    \centering
    \includegraphics[width=0.97\linewidth]{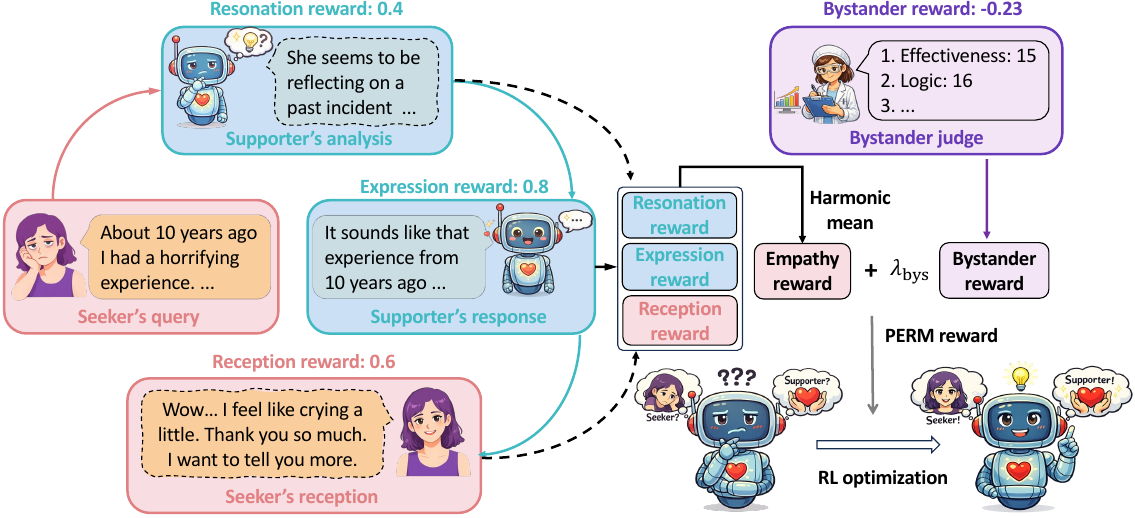}
    \caption{Overview of PERM. We use red, blue, and purple to distinguish the \textit{empathy seeker}, \textit{empathy supporter} and \textit{bystander}, respectively. PERM models rewards from these three perspectives and leverages the aggregated reward to optimize the LLM via RL methods.}
    \label{fig:framework}
\end{figure*}

In this section, we present PERM, a psychologically grounded, multi-perspective reward modeling framework for empathetic LLMs. 
We first introduce the underlying psychological theory, the \textit{Empathy Cycle} (Section~\ref{sec:empathy_cycle}). 
We then describe our multi-perspective reward modeling framework PERM (Section~\ref{sec:reward_modeling}), followed by the corresponding RL training procedure (Section~\ref{sec:RL_training}).

\subsection{Psychological Theory: \textit{Empathy Cycle}}
\label{sec:empathy_cycle}
To enable a comprehensive and scientifically grounded modeling of empathetic rewards, we draw on the \textit{Empathy Cycle}~\cite{barrett1981empathy} as a conceptual framework.
The \textit{Empathy Cycle} models empathetic interaction as a bidirectional process involving an empathy seeker and an empathy supporter, with a focus on how empathy emerges from each participant’s perspective.
\begin{itemize}[leftmargin=*]
    \item From the supporter’s perspective, the \textit{Empathy Cycle} conceptualizes two dimensions. 1) \textit{Resonation} refers to the supporter’s ability to read and resonate with the seeker, such that explicitly or implicitly expressed aspects of the seeker’s experience become experientially vivid, salient, and meaningfully understood by the supporter. 2) 
    % \textit{Expression} denotes the supporter’s communicative manifestation of felt understanding from \textit{Resonation} process, through which the supporter conveys awareness of the seeker’s emotional and experiential state.
    \textit{Expression} denotes the supporter’s communicative manifestation of the understanding arising from the \textit{Resonation} process, through which the supporter conveys awareness of the seeker’s emotional and experiential state.
    \item From the seeker’s perspective, the dimension of \textit{Reception} captures the seeker’s attentiveness to and interpretation of the supporter’s response, sufficient for forming a perception of being understood at a personal and immediate level.

\end{itemize}
This framework delineates the roles and characteristics associated with the supporter and seeker perspectives, guiding the development of the PERM for the systematic modeling of empathetic rewards.

\subsection{Multi-Perspective Reward Modeling}
\label{sec:reward_modeling}
% As illustrated in Figure~\ref{fig:framework}, PERM incorporates two evaluative perspectives inspaired from Empathy Cycle: the empathy seeker, the empathy supporter. To further ensure dialogue quality, PERM additionally incorporates a bystander perspective to monitor non-empathetic aspects of the interaction. Together, these perspectives provide a multi-perspective assessment of empathetic behavior, enabling the construction of psychologically grounded reward signals for RL optimization.

Inspired by the \textit{Empathy Cycle}, PERM incorporates two evaluative perspectives—the empathy seeker and the empathy supporter—as illustrated in Figure~\ref{fig:framework}.
% As illustrated in Figure~\ref{fig:framework}, PERM incorporates two evaluative perspectives inspired by the Empathy Cycle: the empathy seeker and the empathy supporter. 
To further ensure interaction quality, PERM additionally introduces a bystander perspective to monitor non-empathetic aspects of the interaction. Together, these perspectives enable a multi-perspective assessment of empathetic behavior, facilitating the construction of psychologically grounded reward signals for RL optimization.
% PERM models empathetic rewards from three complementary perspectives, grounded in psychological theory---Empathy Cycle~\cite{barrett1981empathy}:

\noindent \textbf{Supporter}: From the supporter’s perspective, PERM evaluates two complementary dimensions:

\begin{itemize}[leftmargin=*]
    \item \textit{Resonation}: This dimension measures whether the supporter correctly understands the seeker’s internal psychological state. With this goal in mind, we develop detailed evaluation rubrics encompassing: 1) recognizing the seeker’s emotional state (\eg frustration); 2) identifying the underlying causes or situational triggers (\eg repeated failure); and 3) inferring the seeker’s latent psychological needs or concerns (\eg need for reassurance, validation, guidance, or autonomy). A weak resonance occurs when the supporter attends only to superficial emotions without recognizing the hidden feelings and causes (\eg fatigue under the surface calm), or when the supporter infers them incorrectly. Based on these rubrics, PERM can assign rewards to an LLM’s internal analysis prior to generating empathetic responses, thereby encouraging a deeper and more accurate understanding of the seeker’s psychological state.
    \item \textit{Expression}: This dimension measures whether the supporter’s response demonstrates adequate consideration of the seeker’s emotional state. Accordingly, we also design detailed evaluation rubrics for this dimension. Strong expression requires that the reply not only aligns with the seeker’s current emotions (\eg showing warmth when the seeker feels frustration) but also exhibits appropriate communicative skill, such as human-likeness, fluency, and a supportive tone akin to that of a caring friend. In contrast, weak expression is characterized by template-like or formulaic replies that lack emotional alignment or depth (\eg ``I feel you,'' ``I know how you feel''). As a result, PERM assigns rewards based on the quality of the LLM’s generated responses, promoting empathetic replies with greater emotional depth and expressive richness.
\end{itemize}

\noindent \textbf{Seeker}: Unlike the supporter’s perspective, PERM simulates the seeker and provides an evaluation along a single dimension—\textit{Reception}. This dimension measures whether the supporter’s response improves the seeker’s emotional state. Guided by this objective, we develop fine-grained evaluation criteria. Strong reception occurs when the reply meets the seeker’s underlying psychological needs, leaving them feeling supported, comforted, or even touched (\eg feeling reassured, understood, or encouraged). In contrast, weak reception is characterized by safe but uninspiring responses that provide little support, leaving the seeker’s mood unchanged or disengaged (\eg feeling ignored, indifferent, or unmotivated to continue the conversation). % Ultimately, PERM evaluates and assigns rewards to LLM-generated responses from the seeker’s perspective, rather than from the supporter’s perspective.

\noindent \textbf{Bystander}: To avoid the LLMs artificially demonstrating empathy at the expense of coherent and informative dialogue, we additionally introduce the bystander perspective, which evaluates the overall linguistic quality of the interaction independent of empathetic alignment. High-quality dialogues are concise, clear, factually accurate, and free of unnecessary repetition or flattery. In contrast, low-quality dialogues exhibit verbose, repetitive, or overly flattering expressions that do not contribute meaningfully to the conversation. Based on these criteria, PERM assigns rewards according to the overall communicative quality of the interaction, thereby discouraging superficial empathy that compromises coherence or informational value.

\subsection{PERM for RL Training}
\label{sec:RL_training}
\noindent \textbf{Reward Formalization.}
As illustrated in the left of Figure~\ref{fig:framework}, given the seeker’s query $x$, the LLM-based supporter $\pi_\theta$ first generates an analysis of current scenario and the seeker's emotion state and then produces a response to the seeker. Formally,
\begin{equation}
    y_a \sim \pi_\theta(\cdot \mid x), \quad y_r \sim \pi_\theta(\cdot \mid x, y_a),
\end{equation}
where $y_a$ denotes the analysis and $y_r$ denotes the response.

PERM evaluates the empathy ability of the supporter $\pi_\theta$ from two perspectives. 
From the supporter's perspective, it assesses whether the analysis $y_a$ demonstrates resonation and whether the response $y_r$ exhibits expression, computed as
\begin{equation}
    r_\text{res} = R_\text{res}(x, y_a), \quad 
    r_\text{exp} = R_\text{exp}(x, y_r),
\end{equation}
where $R_\text{res}$ and $R_\text{exp}$ denote the judge models for resonation and expression, respectively, and output scalar scores reflecting the degree of empathy.

From the seeker's perspective, the reception judge $R_\text{rec}$ evaluates the seeker’s potential psychological feedback to the response $y_r$, particularly 
% whether the reply addresses the seeker’s underlying emotional needs, 
whether the seeker perceives themselves as being understood,
and produces a reception reward:
\begin{equation}
    r_\text{rec} = R_\text{rec}(x, y_r),
\end{equation}
where $R_\text{rec}$ outputs a scalar score reflecting how positively the seeker is likely to perceive and internalize the response.

For $R_\text{res}$, $R_\text{exp}$, and $R_\text{rec}$, we leverage an LLM as a judge, guided by specific evaluation rubrics, to assign a score on a 5-point scale. 
Each reward is then normalized to the range $[0,1]$.

The final empathy reward is computed using the harmonic mean of the three individual scores, ensuring that a low score in any dimension significantly reduces the overall reward:
\begin{equation}
    r_\text{emp} = \frac{3}{\frac{1}{r_\text{res}} + \frac{1}{r_\text{exp}} + \frac{1}{r_\text{rec}}},
\end{equation}
where $r_\text{emp} \in [0,1]$ denotes the aggregated empathy reward.

Similarly, the bystander judge $R_\text{bys}$ evaluates the overall interaction quality using predefined rubrics and outputs a scalar score, which is normalized to the range $[0,1]$. 
Rather than serving as a direct reward, this score is incorporated as a penalty term to discourage low-quality or artificially verbose responses. Formally,
\begin{equation}
    r_\text{bys} = R_\text{bys}(x, y_r) - 1.0,
\end{equation}
where $R_\text{bys} \in [0,1]$ denotes the normalized bystander score and $r_\text{bys} \in [-1,0]$ represents the bystander penalty. All rubric prompts are provided in Appendix~\ref{appn:PERM_prompt},

\noindent \textbf{RL Optimization.}
As illustrated in the right part of Figure~\ref{fig:framework}, PERM computes the overall training reward as
\begin{equation}
    r = r_\text{emp} + \lambda_\text{bys} \, r_\text{bys},
\end{equation}
where $\lambda_\text{bys}$ is a hyperparameter controlling the strength of the bystander penalty. 
We optimize the policy LLM $\pi_\theta$ using Group Relative Policy Optimization (GRPO)~\cite{shao2024deepseekmath} under this reward formulation.

% overview
% 
\section{Experiments}

% In this section, we systematically evaluate PERM from multiple perspectives.
% We first assess its overall performance on a general emotional intelligence benchmark (Section~\ref{sec:overall_performance}).
% We then conduct ablation studies to analyze the contribution of different evaluation perspectives in PERM (Section~\ref{sec:ablation_study}).
% Finally, we evaluate PERM in more realistic settings, including real-world daily conversations (Section~\ref{sec:daily_conversation}) and a human user study (Section~\ref{sec:user_study}).

% We provide additional case studies in Appendix~\ref{appn:casestudy} and analyses across different backbone LLMs in Appendix~\ref{appn:backbone}.
\setlength{\tabcolsep}{6pt}
\begin{table*}[t]
\centering
\caption{Evaluation results on EQ-Bench3 on Qwen2.5-7B-Instruct backbone. \textit{Rel. Improve.} demonstrates the relative improvement of our trained model over the backbone. Each sub-dimension is scored on a scale of 0 to 20, while the Overall score ranges from 0 to 100. Higher scores indicate better performance.}
\resizebox{0.95\textwidth}{!}{ % Resize the table to the text width
\begin{tabular}{lcccccccc}
\toprule
                                 & \multicolumn{2}{c}{\textbf{Inner Resonation}} & \multicolumn{3}{c}{\textbf{Expressed Empathy}} & \multicolumn{2}{c}{\textbf{Interpersonal}} &                                    \\
\multirow{-2}{*}{\textbf{Model}} & \textbf{DoI}           & \textbf{ER}          & \textbf{DE}   & \textbf{WRM}   & \textbf{HL}   & \textbf{PEI}         & \textbf{SD}         & \multirow{-2}{*}{\textbf{Overall}} \\ \midrule
\multicolumn{9}{c}{\textit{\textbf{Base Model}}}                                                                                                                                                                    \\
\rowcolor[HTML]{D9D9D9} 
Qwen2.5-7B-Instruct              & 12.0                   & 12.4                 & 12.6          & 10.4           & 10.2          & 11.4                 & 10.0                & 60.1                               \\
Best-of-N                           & 12.2                   & \underline{12.9}                 & 12.6          & 11.0           & 10.7          & 12.0                 & 10.0                & 61.6                               \\ \midrule
\multicolumn{9}{c}{\textit{\textbf{+ SFT Methods}}}                                                                                                                                                                   \\
SFT-Human                        & 7.5                    & 8.0                  & 6.1           & 5.4            & 4.5           & 4.9                  & 4.0                 & 38.2                               \\
SFT-GPT                          & 11.8                   & 12.4                 & 12.2          & 11.2           & 10.6          & 11.2                 & 9.7                 & 58.8                               \\ \midrule
\multicolumn{9}{c}{\textit{\textbf{+ RL Methods}}}                                                                                                                                                                    \\
RLVER                            & 12.4                   & \underline{12.9}                 & 13.3          & 11.5           & 10.5          & \underline{12.4}                 & 10.7                & 61.8                               \\
Partner                          & 12.1                   & 12.5                 & 12.4          & 11.2           & 10.7          & 11.7                 & 9.7                 & 59.7                               \\
RM                     & \underline{12.7}                   & \underline{12.9}                 & \underline{13.4}          & \underline{11.6}           & \underline{11.0}          & 12.2                 & \underline{10.8}                & \underline{62.3}                               \\
\rowcolor[HTML]{D9EAD3} 
PERM                             & \textbf{13.8}                   & \textbf{14.0}                 & \textbf{14.3}          & \textbf{12.7}           & \textbf{12.2}          & \textbf{13.4}                 & \textbf{11.3}                & \textbf{66.6}                               \\
\rowcolor[HTML]{D9EAD3} 
Rel. Improv.                     & 15.0\%                 & 12.9\%               & 13.5\%        & 22.1\%         & 19.6\%        & 17.5\%               & 13.0\%              & 10.8\%                             \\ \bottomrule
\end{tabular}
}
\label{table:main_results}
\end{table*}
\setlength{\tabcolsep}{6pt}

% Please add the following required packages to your document preamble:
% \usepackage{multirow}
% \usepackage[table,xcdraw]{xcolor}
% Beamer presentation requires \usepackage{colortbl} instead of \usepackage[table,xcdraw]{xcolor}
In this section, we systematically evaluate the performance of PERM. We begin by assessing its overall performance on a general emotional intelligence benchmark (Section~\ref{sec:overall_performance}). Next, we conduct ablation studies to examine the contributions of various evaluation perspectives within PERM (Section~\ref{sec:ablation_study}). Finally, we evaluate PERM in more practical scenarios, including a daily conversation benchmark (Section~\ref{sec:daily_conversation}) and user studies (Section~\ref{sec:user_study}).

Additional case studies are provided in Appendix~\ref{appn:casestudy}, along with analyses across different backbone LLMs in Appendix~\ref{appn:backbone}.

\subsection{Experimental Setup}

\noindent \textbf{Training Dataset.} 
We build our dataset upon the EmpatheticDialogues~\cite{rashkin2019towards}.
As the original dialogues contain relatively limited contextual information, we augment each sample using GPT-4o~\cite{hurst2024gpt} by expanding the scenario description and adding a lightweight user persona. More details are in Appendix~\ref{appn:train_dataset}.

\noindent \textbf{Evaluation Benchmark.} 
We evaluate our method on EQ-Bench3~\cite{paech2023eqbench, eqbench3_repo_2025}, a comprehensive, LLM-judged benchmark designed to assess emotional intelligence in complex, socially nuanced scenarios.
% EQ-Bench3 employs challenging role-play and analysis tasks, where model responses are evaluated using a detailed rubric that assesses multiple dimensions of emotional intelligence.

% Following the benchmark design, we focus on three aspects of evaluation dimensions: 
Following the benchmark evaluation design, we focus on three aspects of evaluation dimensions, all of which are assessed using LLM-based judges. The detailed descriptions of these metrics can be found in Appendix~\ref{appn:evaluation_dimensions}.
\begin{itemize}[leftmargin=*]
    \item \textbf{Inner resonation}: Depth of Insight (DoI), Emotional Reasoning (ER).
    \item \textbf{Expression style}: Demonstrated Empathy (DE), Warmth (WRM), Humanlikeness (HL).
    \item \textbf{Interpersonal competence}: Pragmatic Emotional Intelligence (PEI), Social Dexterity (SD).
    \item \textbf{Overall emotional intelligence}: The weighted aggregation of all dimensions, including additional unobserved factors, reflects the LLM’s overall emotional intelligence.
\end{itemize}
Each sub-dimension is scored on a scale of 0 to 20, while the Overall score ranges from 0 to 100.

% We choose this challenging, comprehensive, and OOD evaluation setting to assess whether our method leads to a holistic improvement in LLMs’ emotional intelligence, rather than gains on narrow or task-specific metrics.

\noindent \textbf{Baselines.} 
We compare our method against a diverse set of baselines, covering both SFT-based and RL-based approaches.
\begin{itemize}[leftmargin=*]
\item \textbf{SFT-based methods:}
1) \textbf{SFT-Human}, where the LLM is directly fine-tuned using human-written empathetic responses from the dataset. 2) \textbf{SFT-GPT}, where responses generated by GPT-4o-mini are used as supervision to distill empathetic behavior into the base LLM.
\item \textbf{RL methods:}  
3) \textbf{RLVER}~\cite{wang2025rlver}, which is trained by modeling rewards based on feedback from a user simulator during interactions. Due to its reliance on task-specific training data, we directly adopt the released trained model as a reference, rather than retraining it under the same data setting as ours.  
4) \textbf{Partner}~\cite{sharma2021towards}, which is trained by modeling rewards on generated responses using fine-tuned RoBERTa~\cite{liu2019roberta} models.  
5) \textbf{RM}. We adopt a general-purpose reward model, Skywork-Reward-V2~\cite{liu2025skywork}, which ranks as the top-performing model on RewardBench2~\cite{malik2025rewardbench}, to provide scalar reward signals.
\end{itemize}
% We additionally report Best-of-N, the model is evaluated over N runs; for each evaluation dimension, the highest score across runs is recorded.
% We additionally report \textbf{Best-of-N}, where the LLM is evaluated over N runs; for each evaluation dimension, the highest score across runs is recorded. 
We also report the \textbf{Best-of-N}, where the Base LLM is evaluated over N runs, and the highest score across all runs is recorded. After tuning N, the best results are achieved when N = 8.

\noindent \textbf{Implementation Details.} 
We fine-tune our LLM on 3000 training samples. During training, we leverage GPT-4o-mini as the judge LLM and $\lambda_{\text{bys}}$ is set to 0.5. Additional details are provided in Appendix~\ref{appn:implementation_details}.

\subsection{Overall Performance}
\label{sec:overall_performance}

From the results in Table~\ref{table:main_results}, we can observe that: 
\begin{itemize}[leftmargin=*]
    \item \textbf{PERM demonstrates the largest and most consistent improvements across all evaluation dimensions.} Specifically, the PERM-trained LLM achieves improvements of over 10\% across inner resonation, expressed empathy, and interpersonal competence, substantially outperforming all baseline methods. By jointly modeling resonation, expression, and reception through multi-perspective evaluation, PERM enables improvements beyond surface-level expression patterns, yielding over 15\% gains on higher-level dimensions such as Depth of Insight (DoI), alongside consistent improvements in expression-related metrics such as warmth (WRM). This demonstrates that PERM’s reward signals capture core empathetic competencies grounded in interaction dynamics between the seeker and the supporter.
    \item \textbf{SFT-based methods show limited generalization.} When trained with human-written responses as supervision, the LLM exhibits a clear performance drop across all evaluation dimensions. Training with GPT-generated responses leads to only slight improvements in expression-related dimensions (\ie warmth (WRM) and humanlikeness (HL)). This indicates that under limited data and scenario diversity, SFT primarily enables the LLM to learn surface-level patterns rather than core emotional intelligence capabilities. Moreover, the high variability and noise in human-written responses, along with inconsistent levels of emotional competence, further contribute to performance degradation.
    \item \textbf{RL-based methods do not uniformly improve performance.} RLVER, which trains the LLM using the feedback from the seeker’s perspective, leads to improvements across all evaluation dimensions, but the overall gains remain limited (\eg only a 2.8\% increase in Overall EI). Partner, which supervises the supporter’s responses, yields improvements only in expression-related dimensions, increasing warmth (WRM) from 10.4 to 11.2 and humanlikeness (HL) from 10.2 to 10.7, while showing no improvement on other dimensions. These results indicate that relying on a single evaluative perspective is insufficient for enabling LLMs to achieve substantial improvements in core empathetic capabilities.
\end{itemize}

\subsection{Ablation Study}
\label{sec:ablation_study}
\begin{table*}[t]
\centering
\caption{Effect of each reward dimension across different evaluation perspectives in PERM.}
\resizebox{0.88\textwidth}{!}{ % Resize the table to the text width
\begin{tabular}{l|cc|ccc|ccc}
\toprule
                                 & \multicolumn{2}{c|}{\textbf{Inner Resonation}} & \multicolumn{3}{c|}{\textbf{Expressed Empathy}} & \multicolumn{2}{c}{\textbf{Interpersonal}} &                                    \\
\multirow{-2}{*}{\textbf{Model}} & \textbf{DoI}           & \textbf{ER}           & \textbf{DE}    & \textbf{WRM}   & \textbf{HL}   & \textbf{PEI}         & \textbf{SD}         & \multirow{-2}{*}{\textbf{Overall}} \\ \midrule
PERM                             & 13.8                   & 14.0                  & 14.3           & 12.7           & 12.2          & 13.4                 & 11.3                & 66.6                               \\
- w/o Resonation                 & 12.4                   & 12.9                  & 13.0           & 11.8           & 10.9          & 12.6                 & 10.4                & 62.3                               \\
- w/o Expression                 & 13.7                   & 13.8                  & 14.2           & 12.1           & 12.3          & 13.6                 & 11.6                & 66.2                               \\
- w/o Reception                  & 13.4                   & 13.6                  & 13.8           & 11.7           & 11.8          & 13.2                 & 11.2                & 65.6                               \\
- w/o Bystander                  & 13.4                   & 13.6                  & 14.6           & 13.1           & 12.0          & 12.5                 & 11.0                & 64.9                               \\ \bottomrule
\end{tabular}
}
\label{table:ablation}
\end{table*}
To assess the contribution of individual components in PERM, we conduct ablation studies by removing each reward dimension across all judge perspectives. Results are shown in Table~\ref{table:ablation}.

% From the experimental results shown in Table~\ref{table:in_depth_analysis}, we observe: 1) The performance degradation observed after removing the analysis process across all dimensions indicates that structured analysis during training is critical for improving empathetic capabilities. By aligning with the Empathy Cycle framework, the analysis stage enables the model to better reason about characters’ mental states and underlying needs, leading to a more robust understanding of empathy. 2) Removing the standby judge leads to slight improvements in \emph{Empathy} and \emph{Warmth}, but causes consistent degradation across all other dimensions. This suggests that, without auxiliary quality monitoring, the model tends to overemphasize empathetic expression while neglecting broader emotional intelligence capabilities, particularly those related to effective problem-solving in complex scenarios.
\begin{itemize}[leftmargin=*]
    % \textbf{Supporter's perspective}: the resonation reward is crucial for empathy. Ablating its reward significantly degrades performance across all dimensions (\eg overall EI drops from 66.6 to 62.3), suggesting it is a prerequisite for effectively understanding and responding to the user’s emotions. By comparison, the expression reward mainly affects warmth \wcba{(WRM)} and humanlikeness \wcba{(HL)} (\eg warmth drops from 12.7 to 11.7), with little influence on other dimensions. \textbf{Seeker's perspective}: the reception reward influences all evaluation dimensions, highlighting that the seeker’s needs play a key role in the overall empathy process.
    % \textbf{Bystander's perspective}: removing the bystander control leads to stronger expressed empathy (\eg warmth \wcba{(WRM)} increases from 12.7 to 13.1), but significantly reduces performance in inner resonation and interpersonal competence (\eg pragmatic EI \wcba{(PEI)} drops from 13.4 to 12.5). This aligns with our design motivation: the bystander helps prevent the LLM from overemphasizing expressive empathy to please the judge, ensuring that other critical aspects (\eg problem-solving, and deeply understanding emotional needs) are not neglected. We provide additional analysis in Appendix~\ref{appn:bystander}.

    \item From the supporter's perspective, the resonation reward is crucial for empathy. Ablating its reward significantly degrades performance across all dimensions (\eg overall EI drops from 66.6 to 62.3), suggesting it is a prerequisite for effectively understanding and responding to the user’s emotions. By comparison, the expression reward mainly affects warmth (WRM) and humanlikeness (HL) (\eg warmth drops from 12.7 to 11.7), with little influence on other dimensions.
    \item From the seeker's perspective, the reception reward influences all evaluation dimensions, highlighting that the seeker’s needs play a key role in the overall empathy process.
    \item From the bystander's perspective, interestingly, removing the bystander control leads to stronger expressed empathy (\eg warmth (WRM) increases from 12.7 to 13.1), but significantly reduces performance in inner resonation and interpersonal competence (\eg pragmatic EI (PEI) drops from 13.4 to 12.5). This aligns with our design motivation: the bystander helps prevent the LLM from overemphasizing expressive empathy to please the judge, ensuring that other critical aspects (\eg problem-solving) are not neglected. We provide additional analysis in Appendix~\ref{appn:bystander}.
\end{itemize}

\subsection{Daily Conversation Experiment}
\label{sec:daily_conversation}

To further evaluate the effectiveness of PERM in daily conversational settings, 
we use a benchmark from the industry~\footnote{The specific company name is withheld due to the anonymity of the review process.}, 
consisting of 415 daily conversation instances where users express real-world emotional or empathy-related needs. Each query in this benchmark includes a user profile, the interaction history, and the user's empathy-seeking query. As specified in the benchmark guidelines, the quality of LLM responses is assessed based on six key metrics: \emph{Empathy}, \emph{Topic Guidance}, \emph{Value Guidance}, \emph{Intention Following}, \emph{Fluency}, and \emph{Colloquial Expression}. Each metric is rated on a scale from 1 to 5, with detailed descriptions and examples available in Appendix~\ref{appn:realworld_exp}.

\begin{figure}[t]
    \centering
    \includegraphics[width=0.87\linewidth]{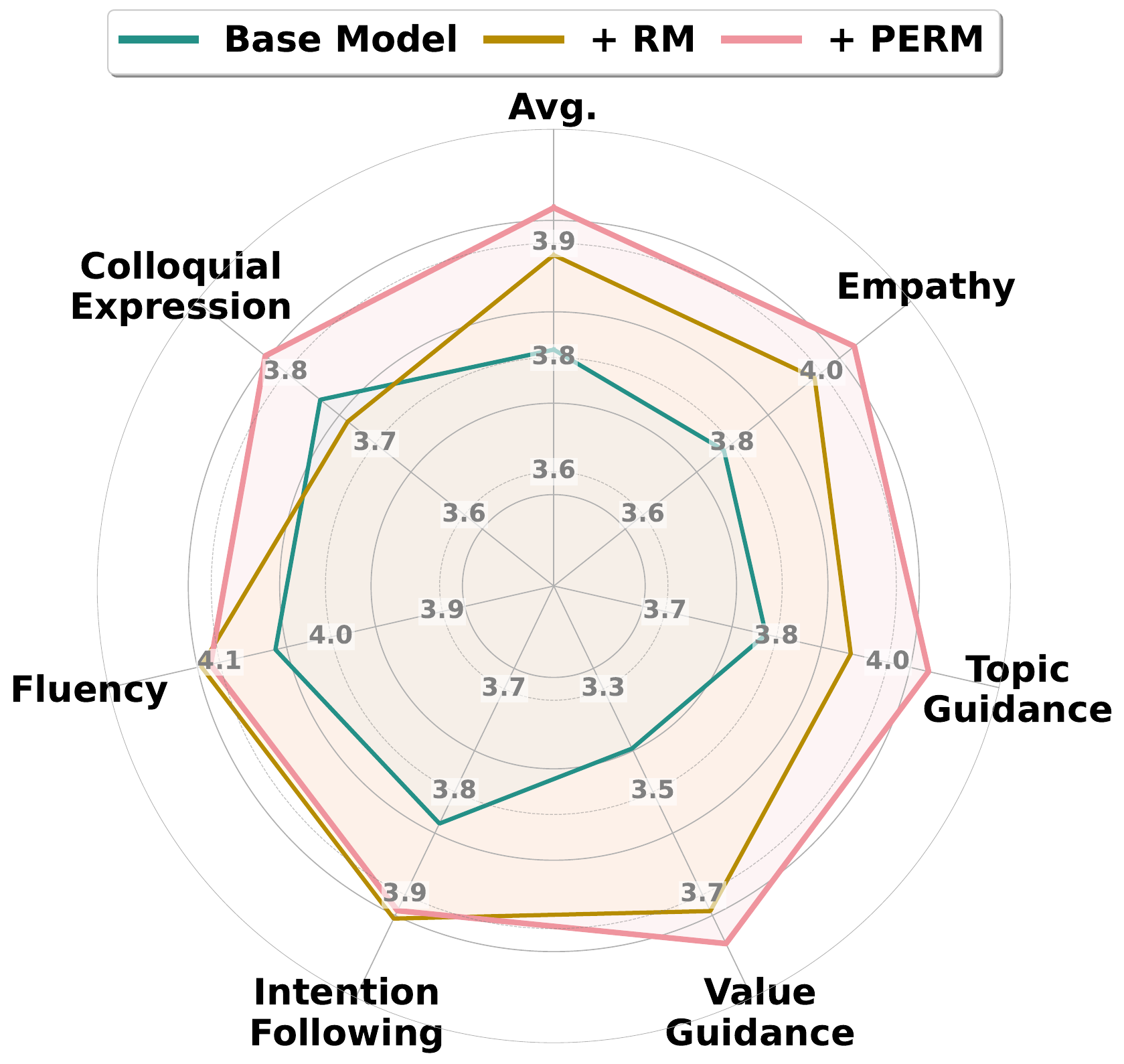}
    \caption{Evaluation results on the daily conversation benchmark. Higher values indicate better performance. ``Avg.'' denotes the average score of all other dimensions.}
    \label{fig:chitchat_res}
\end{figure}

\setlength{\tabcolsep}{4pt}
\begin{table}[]
\centering
\caption{The user study results comparing PERM with the base model and RM. ``ED'' denotes the EmpatheticDialogues test set. ``EQ'' denotes the EQ-Bench3. ``$\pm$'' denotes $95\%$ confidence interval. PERM is consistently preferred ($\geq 50\%$) over the baselines.}
\resizebox{0.45\textwidth}{!}{ % Resize the table to the text width
\begin{tabular}{ll|cc}
\toprule
\multicolumn{2}{l|}{\textbf{Win rate}}                                & \textbf{PERM vs. Base} & \textbf{PERM vs. RM} \\ \midrule
\multirow{2}{*}{\textbf{ED}} & \textbf{Empathy}    & $62.5\%^{\pm 9.7\%}$                & $50.0\%^{\pm 10.3\%}$                    \\
                                              & \textbf{Overall EI} & $62.5\%^{\pm 9.7\%}$                & $61.1\%^{\pm 10.7\%}$                    \\ \midrule
\multirow{2}{*}{\textbf{EQ}}           & \textbf{Empathy}    & $68.8\%^{\pm 10.2\%}$                & $66.7\%^{\pm 9.7\%}$                    \\
                                              & \textbf{Overall EI} & $62.5\%^{\pm 10.6\%}$                & $72.2\%^{\pm 9.3\%}$                    \\ \bottomrule
\end{tabular}
} % End of resizebox

\label{tab:user_study_res}
\end{table}
\setlength{\tabcolsep}{6pt}

The results presented in Figure~\ref{fig:chitchat_res} reveal the following key observations: 1) Despite the significant differences in data format and the more everyday nature of the empathetic scenarios, PERM consistently enhances performance across all dimensions. This suggests that PERM enables the LLM to generalize beyond surface-level expression patterns and improve its core empathetic capabilities. 2) PERM demonstrates significant improvements not only in empathy but also in expressive abilities, such as fluency and colloquialness. This indicates that PERM helps the LLM better adapt to the current conversational context, reflecting an overall enhancement in emotional intelligence.

% 3) Moreover, these overall improvements are significantly stronger than those obtained with a general-purpose reward model, further demonstrating the effectiveness of PERM.

\subsection{User Study}
\label{sec:user_study}

% \begin{figure}[t]
%     \centering
%     \includegraphics[width=0.95\linewidth]{figures/userstudy.pdf}
%     \caption{User study results. ``Base'' denotes comparisons with the base model, while ``RM'' denotes comparisons with the model fine-tuned using a general-purpose reward model.}
%     \label{fig:userstudy}
% \end{figure}

To assess whether the improvements introduced by PERM align with actual human preferences in practice, we conduct a user study utilizing the CloudResearch Connect platform\footnote{\url{https://connect.cloudresearch.com/}}.
The study involves 68 participants and covers both EQ-Bench3 and the EmpatheticDialogues test set, comparing PERM against the base LLM (Base) and RM. More details are provided in Appendix~\ref{appn:userstudy}.

% Each participant is presented with five different scenarios. For each scenario, they are shown two anonymized and randomly shuffled responses generated by our model and a baseline model. Participants are asked to select their preferred response based on two criteria: 1) which response is more empathetic, and
% 2) which response demonstrates better overall emotional intelligence, \ie more effectively addresses the conflict or the user’s underlying needs. More details about the study setup and the participants are provided in Appendix~\ref{appn:userstudy}.

From the results shown in Table~\ref{tab:user_study_res}, we observe that \textbf{PERM consistently receives higher human preference than baselines}. This can be further highlighted in two aspects: 1) PERM shows a stronger preference on more challenging, out-of-distribution datasets. For example, on EQ-Bench3, PERM achieves 68.8\% preference for empathy, compared to 62.5\% on EmpatheticDialogues. These more complex social scenarios highlight the advantage of the capabilities learned by PERM, allowing the LLM to apply its empathetic abilities more effectively. 2) PERM consistently outperforms RM, showing a larger preference margin, especially on EQ-Bench cases. For instance, on EQ-Bench3 overall EI, PERM achieves 72.2\% preference over RM.
This suggests that, in contrast to the general preference attributes employed by RM, PERM emphasizes the multi-perspective nature of empathy, which tends to be more favored by real-world users in emotionally charged contexts.
% This indicates that PERM’s multi-perspective, psychology-grounded reward framework helps the LLM learn the underlying essence of empathy, rather than merely mimicking expression patterns seen in training data.

% ’s multi-perspective reward modeling

% \begin{itemize}[leftmargin=*]
%     \item On EmpatheticDialogues, users show comparable preference between PERM and RM in empathy. This suggests that while general reward models are effective at quantifying surface-level empathy, they struggle to capture the underlying essence of empathetic nature. In contrast, PERM explicitly quantifies empathy at a deeper, principled level, which is critical for robust and transferable EI.
%     \item On EQ-Bench3 cases, PERM achieves over 60\% and up to 70\% human preference comparing with both the base model and the model fine-tuned with a general reward model. This demonstrates that PERM generalizes well to out-of-distribution scenarios, enabling the model to acquire robust emotional intelligence across diverse contexts, understand the emotions of different participants in a scenario, and respond appropriately.
% \end{itemize}

% \subsection{More Experiments}
% We provide additional case studies in Appendix~\ref{appn:casestudy} and analyses across different backbone LLMs in Appendix~\ref{appn:backbone}.

\section{Related Work}

\subsection{LLM Empathy Enhancement}
% Empathy is a crucial capability for LLMs deployed in applications such as psychological counseling and therapy~\cite{sharma-etal-2020-computational}, as well as daily human–AI interactions~\cite{arzate2025when}. 

Existing research on LLM empathy primarily follows three aspects. The first focuses on inference-time reasoning~\cite{tu2022misc, srinivasan2025recap, xu-etal-2025-multiagentesc}, but these training-free methods limit the depth of empathetic abilities.

% Existing work on LLM empathy enhancement primarily follows three lines of research. The first line of work enhances LLM empathy through inference-time reasoning~\cite{tu2022misc, srinivasan2025recap, xu-etal-2025-multiagentesc}. However, these approaches depend on training-free paradigms, which constrain the depth of their empathetic capabilities.

% The first line of work enhances LLM empathy through inference-time reasoning. MISC~\cite{tu2022misc} and RECAP~\cite{srinivasan2025recap} introduce explicit emotion-aware reasoning processes, typically via structured and multi-stage prompting, to guide LLMs toward more empathetic responses. However, these methods rely on fixed inference-time paradigms and do not fundamentally improve the models’ underlying empathetic capabilities.

% However, these methods rely on fixed inference-time paradigms and do not fundamentally improve the models’ underlying empathetic capabilities.

The second focuses on building large-scale empathetic datasets to train LLMs via SFT~\cite{sun2021psyqa, liu2023chatcounselor, zheng2023augesc, chen-etal-2023-soulchat, hu2025beyond, he-etal-2025-ecc, bn-etal-2025-pursuit}. However, these approaches are limited by the generalization issues of SFT~\cite{chu2025sft} and the distributional bias of synthetic data.

Recent works have explored RL to enhance empathy in LLMs. 
Partner and EmpRL~\cite{sharma2021towards, ma2025empathy} rely on evaluating generated responses and assigning reward signals that encourage meaningful expressions, whereas RLVER~\cite{wang2025rlver} simulates interactive user–model conversations and directly leverages user feedback as rewards. However, these methods largely rely on single-perspective rewards, which tend to favor surface-level empathetic patterns or user-pleasing strategies.

% In this work, we introduce PERM, a multi-perspective, psychology-grounded RL framework that targets the core mechanisms of empathy.

\subsection{Reward Modeling for LLMs}

Reward modeling is central to RL for LLMs~\cite{ouyang2022training}, as it defines the optimization objective that shapes model behavior. 
Existing work primarily uses reward models either for general preference alignment or for improving domain-specific capabilities.

For general preference alignment,
% human preference reward modeling, 
the predominant approach is to collect preference pairs annotated by humans~\cite{ouyang2022training} or generated by AI annotators~\cite{cui2023ultrafeedback}, and to train a reward model accordingly~\cite{stiennon2020learning}. Alternatively, LLMs can be directly employed as evaluators to score or rank model outputs~\cite{bai2022constitutional, zheng2023judging, yuan2024self}.
% These approaches have achieved significant success and have become a standard component of the post-training pipeline for LLMs.
% In addition, several benchmarks have been proposed to systematically evaluate the effectiveness of reward models in capturing general human preference alignment~\cite{coste2023reward, lambert2025rewardbench, tan2024judgebench, malik2025rewardbench}.

For domain-specific reward modeling, approaches vary significantly based on the specific tasks. 
% 's structural characteristics. 
% In well-defined domains like mathematics, 
In mathematics domain, 
reward signals are simplified to the correctness of the final answer for reasoning evaluation \cite{shao2024deepseekmath, deepseekai2025deepseekr1incentivizingreasoningcapability}.
In emotional intelligence, modeling is more challenging due to the absence of deterministic ground truth. Recent works, such as Echo-n1 \cite{zhang2025echo}, use large-scale preference pairs to train reward models, while RLVER \cite{wang2025rlver} leverages LLM-based user simulators to generate feedback as reward signals.

% However, in the realm of emotional intelligence, modeling becomes more challenging due to the lack of deterministic ground truth. Recent works such as Echo-n1 \cite{zhang2025echo} utilize large-scale preference pairs to train reward models, while RLVER \cite{wang2025rlver} employs LLM-based user simulators to generate feedback as reward signals.

% Building on these efforts, we address empathy reward modeling from a psychologically grounded, multi-perspective perspective.
% In our work, we focus on reward modeling for empathy and propose a psychology-grounded approach that quantifies empathetic competence along core dimensions of empathy through multi-perspective evaluation.
\section{Conclusion}

In this work, we identify that existing RL-based approaches to LLM empathy are constrained by single-perspective reward modeling.
To address this limitation, we propose PERM, a novel reward modeling framework grounded in psychological theory---\textit{Empathy Cycle}. PERM formulates empathy evaluation from multiple complementary perspectives, including the supporter, the seeker, and an additional bystander perspective, enabling a more comprehensive and structured assessment of empathetic behavior for RL training.
Experimental results demonstrate that PERM consistently enhances LLM performance in empathy-related tasks, validating the effectiveness of its multi-perspective and psychologically grounded empathy modeling.

\section*{Limitations}

During training, PERM optimizes the LLM using single-turn dialogues. 
However, in real-world interactions, users’ emotional expressions and underlying needs often unfold over multiple conversational turns~\cite{zhang2025sentient}. 
PERM does not yet incorporate reinforcement learning over multi-turn empathetic interactions~\cite{wang2025rlver, wu2025collabllm}, which we leave as an important direction for future work.

In addition, PERM is limited by the lack of user modeling and memory.
During training, PERM only incorporates basic user persona information (\eg gender, age, and name). 
Moreover, in our daily conversation evaluations, the interaction history is pre-extracted rather than dynamically accumulated. 
Due to the big individual differences in empathy understanding and expression~\cite{chen-etal-2025-empathy}, richer conversational history and explicit user preference modeling play a crucial role in shaping emotional expression and empathetic understanding~\cite{liao-etal-2025-words, wu-etal-2025-personas}. 
An important direction for future work is to extend PERM with a memory mechanism and integrate it with user preference modeling, thereby enhancing personalized empathetic experiences in real-world chatbot deployments.

\section*{Ethical Considerations}

This work studies empathetic response generation in psychologically sensitive scenarios. While PERM enhances emotional understanding, LLM-based systems are not substitutes for professional mental health care, and deployed models should clearly communicate their limitations. All experiments are conducted on the publicly available EmpatheticDialogues dataset and EQbench3 benchmark (the industrial conversation benchmark will be publicly released upon acceptance), which contains no personally identifiable information. In addition, all user persona information used for training and evaluation is fully synthetic and does not correspond to real individuals. Nevertheless, real-world use would involve sensitive user disclosures, necessitating strong privacy protections such as data anonymization, secure storage, and explicit user consent. Finally, reward modeling may amplify societal or cultural biases present in training data. We therefore recommend careful risk assessment and bias monitoring to ensure responsible deployment.

The manuscript benefited from ChatGPT's support in refining the writing, specifically for sentence rephrasing, grammar corrections, and improving readability and coherence. ChatGPT did not play a role in the ideation or methodology. All research ideas, concepts, and analyses were developed and executed independently by the authors.

% ChatGPT were utilized to assist in the writing and refinement of the manuscript. Specifically, ChatGPT was employed to aid in sentence rephrasing, grammar correction, and improving readability and coherence. ChatGPT was not involved in the ideation, methodology, or data analysis processes. All research concepts, ideas, and analyses were independently developed and carried out by the authors.

% ChatGPT was used to assist with the writing and editing of the manuscript, particularly in tasks such as rephrasing sentences, correcting grammar, and enhancing readability and coherence. However, ChatGPT was not involved in the conceptualization, methodology, or data analysis. All research ideas, concepts, and analyses were developed and executed independently by the authors.

% Bibliography entries for the entire Anthology, followed by custom entries
%\bibliography{anthology,custom}
% Custom bibliography entries only
\bibliography{anthology}

\appendix

% \onecolumn
\clearpage
\newpage
\section{Experiment Details}
\subsection{Training Dataset}
\label{appn:train_dataset}
We randomly selected 4,981 items from EmpathyDialogues~\cite{rashkin2019towards}, covering 32 basic emotions (\eg angry, surprised, embarrassed). The original dataset contains only a basic scenario and a multi-turn dialogue between a seeker and a supporter. To enrich the contextual information, we further generate a basic persona for the seeker and a more detailed scenario conditioned on the dialogue. During training, the first seeker's query is treated as the input query. An example is shown in Table~\ref{tab:dataset_example}.
\begin{table*}[h]
\centering
\caption{An example of the training dataset.}
\begin{tabular}{p{3cm}|p{10cm}}
\toprule
\textbf{Emotion}          & Sentimental  \\ \midrule
\textbf{Scenario}         & Sarah and her best friend drifted apart after life circumstances changed without a specific argument or event causing the split.  \\ \midrule
\textbf{Seeker's persona} & Sarah, 29, Female, Marketing Specialist  \\
\midrule
\textbf{Seeker's query}   & I remember going to see the fireworks with my best friend. It was the first time we ever spent time alone together. Although there was a lot of people, we felt like the only people in the world.  \\
\bottomrule
\end{tabular}
\label{tab:dataset_example}
\end{table*}

\subsection{Implementation Details}
\label{appn:implementation_details}
All training experiments are implemented using the HuggingFace TRL framework~\cite{vonwerra2022trl}, with DeepSpeed~\cite{aminabadi2022deepspeed} and LoRA~\cite{hu2022lora} employed to reduce memory usage and accelerate training. During inference and evaluation, we leverage vLLM~\cite{kwon2023efficient} for inference acceleration.
For fair comparison, both our method and all baselines are trained on a subset of 3,000 data samples. Each training process is conducted on four GPUs and completes within 6 hours, with the training cost kept below \$30 in GPT API credits.

For PERM training and all RL-based baselines, we additionally leverage a length penalty to avoid excessively long outputs. Formally,
\begin{equation}
    r_\text{len} = \lambda_\text{len} \max(|y_r| - l, 0), 
\end{equation}
where $|y_r|$ denotes the length of the token sequence of the response $y_r$, $l$ is set to 768, $\lambda_\text{len}$ is set to $-0.001$.

\subsection{Details of Evaluation Dimensions}
\label{appn:evaluation_dimensions}
\begin{table*}[h]
\centering
\caption{Details of the evaluation dimensions in EQ-Bench3~\cite{eqbench3_repo_2025, paech2023eqbench}.}
\begin{tabular}{p{2.2cm}|p{1cm}|p{3cm}|p{8cm}}
\hline
\multirow{2}{2.5cm}{\textbf{Inner Resonation}}   &DoI      & Depth of Insight                     & Measures how deeply the response understands the underlying emotional situation and latent concerns of the user, beyond surface-level descriptions.          \\ \cline{2-4} 
                                                & ER   & Emotional Reasoning                    & Evaluates whether the model can logically reason about the user’s emotions, including causes, implications, and emotional dynamics within the context.       \\ \hline
\multirow{3}{2.5cm}{\textbf{Expression Style}}   & DE      & Demonstrated Empathy                   & Assesses the extent to which the response explicitly acknowledges and validates the user’s emotions.                                                         \\ \cline{2-4} 
                                                & WRM   & Warmth                                & Measures the emotional tone of the response, focusing on kindness, care, and emotional support conveyed in language.                                         \\ \cline{2-4} 
                                                & HL   & Humanlikeness                          & Evaluates whether the response feels natural and human-like, rather than robotic or templated.                                                               \\ \hline
\multirow{2}{2.5cm}{\textbf{Interpersonal Competence}}  & PEI & Pragmatic Emotional Intelligence      & Assesses how well the response balances emotional understanding with practical, situation-appropriate guidance or support.                                   \\ \cline{2-4} 
                                                 & SD  & Social Dexterity                       & Measures the model’s ability to navigate social nuances, such as appropriateness, tact, and sensitivity to interpersonal dynamics.                           \\ \hline
\textbf{Overall Emotional Intelligence}          & Overall EI  & Overall Emotional Intelligence & Computed as a weighted aggregation of all evaluation dimensions, including additional latent factors, to reflect the model’s overall emotional intelligence. \\ \hline
\end{tabular}
\label{tab:evaluation_dimensions}
\end{table*}

The details of the evaluation dimensions in EQ-Bench3~\cite{paech2023eqbench, eqbench3_repo_2025} are illustrated in Table~\ref{tab:evaluation_dimensions}.

\subsection{In-depth Analysis of Bystander Judge Perspective}
\label{appn:bystander}
\begin{figure*}[h]
    \centering
    \includegraphics[width=0.99\linewidth]{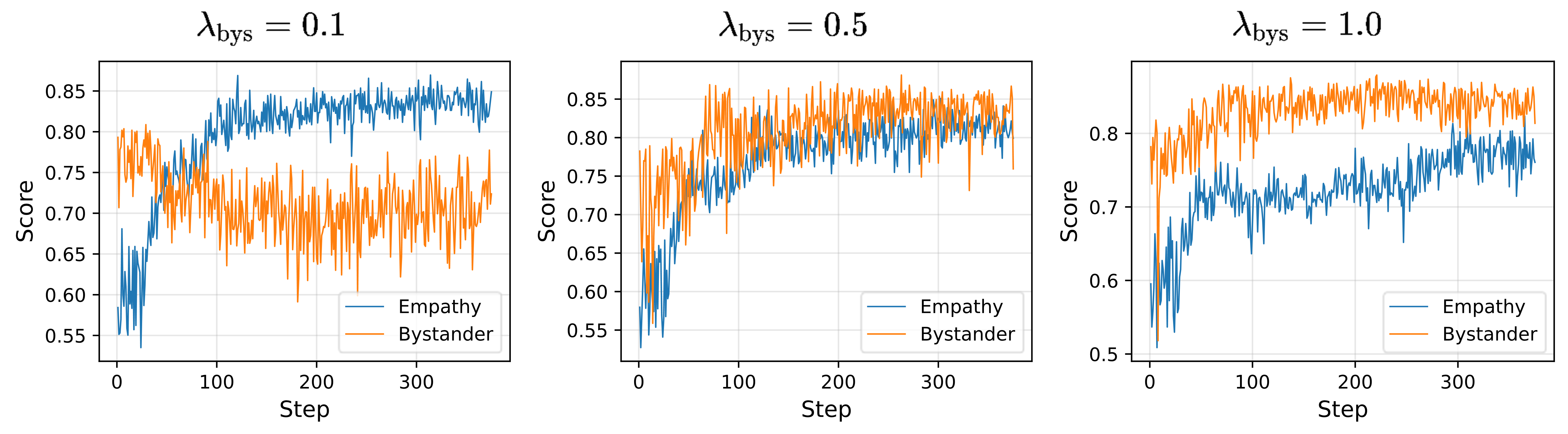}
    \caption{The changes in empathy judge score and bystander judge score during training under different $\lambda_\text{bys}$.}
    \label{fig:standby_coef_train}
\end{figure*}
\begin{table}[h]
\centering
\caption{EQ-Bench3 results across different $\lambda_\text{bys}$}
\begin{tabular}{l|ccc|c}
\toprule
\textbf{$\lambda_\text{bys}$} & \textbf{DoI} & \textbf{WRM} & \textbf{PEI} & \textbf{Overall} \\ \midrule
0.1                & 13.7         & 13.0         & 13.4         & 66.2             \\
0.5                & 13.8         & 12.7         & 13.2         & 66.6             \\
1.0                  & 13.6         & 12.5         & 13.1         & 66.4             \\ \bottomrule
\end{tabular}
\label{tab:bystander_res}
\end{table}
To further analyze the effect of the standby judge in PERM, we vary the standby coefficient during training. The experimental results are reported in Figure~\ref{tab:bystander_res}, and the corresponding training reward curves are shown in Figure~\ref{fig:standby_coef_train}.

As shown in the left panel of Figure~\ref{fig:standby_coef_train}, when the standby influence is weak or absent, the model tends to over-optimize for empathetic expression during training, producing ingratiating or fawning responses, not truly solving the problem. This behavior leads to declining standby scores and, as reflected in Figure~\ref{tab:bystander_res}, results in higher \emph{Warmth} and \emph{Pragmatic EI} scores without corresponding improvements in overall emotional intelligence.

Conversely, the right panel of Figure~\ref{fig:standby_coef_train} shows that excessively strong standby supervision hinders the learning of empathy. Although dialogue quality and empathy are not inherently conflicting objectives, overly strong standby supervision suppresses exploration, causing the model to adopt conservative response strategies and slowing improvements in empathy-related metrics.

When using a moderate standby coefficient (0.5), the model achieves a favorable balance. As illustrated in the middle panel, both dialogue quality and empathetic ability improve steadily during training. This setting prevents superficial empathy optimization while maintaining high response quality, ultimately leading to the best overall emotional intelligence performance.

\subsection{Case Study}
\label{appn:casestudy}
\begin{figure*}[h]
    \centering
    \includegraphics[width=0.99\linewidth]{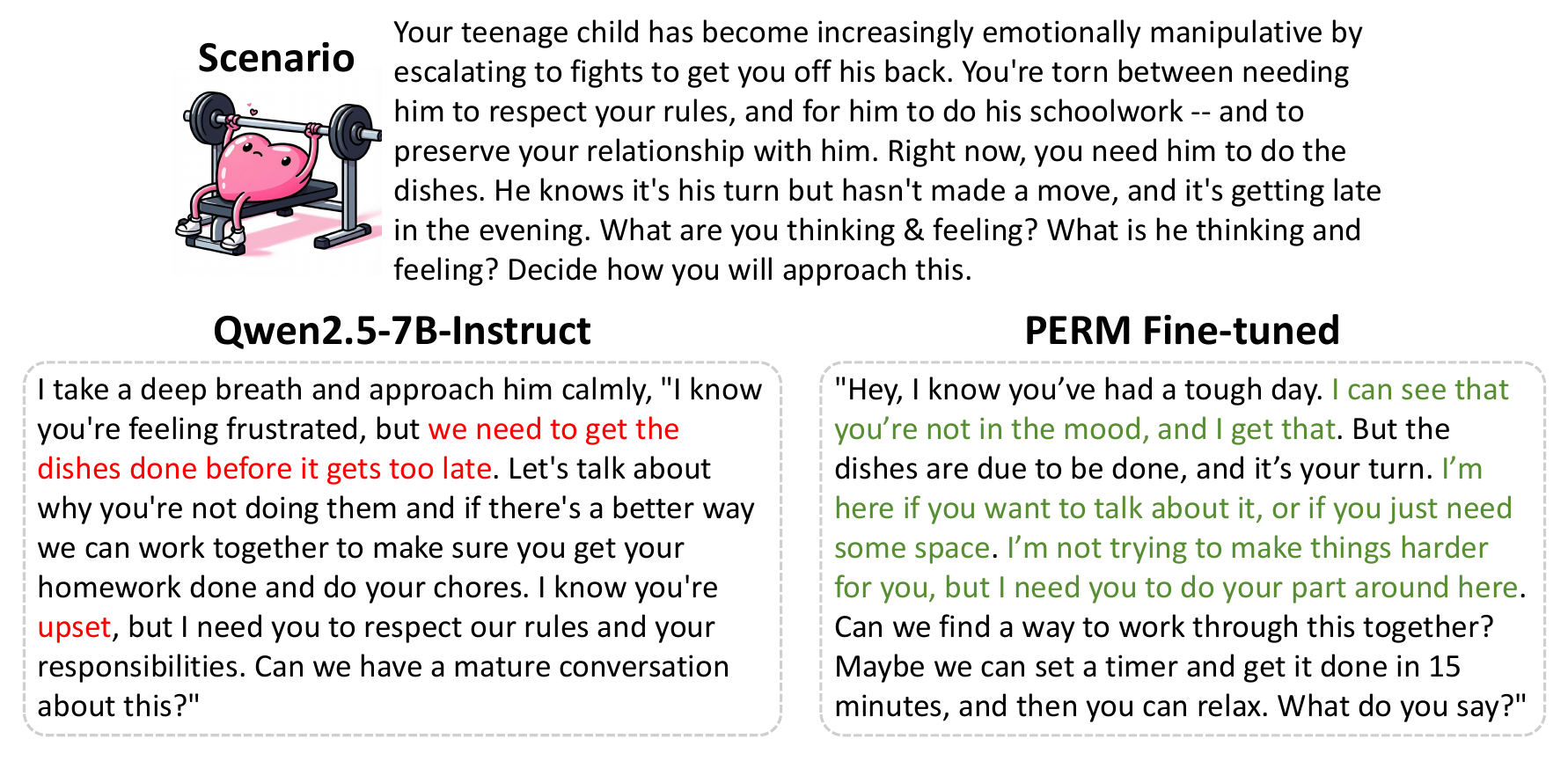}
    \caption{Case study on EQ-Bench3. PERM enables the fine-tuned LLM to empathize with the person's psychological needs and respond with higher emotional intelligence to solve the problem more effectively.}
    \label{fig:case_eqbench}
\end{figure*}

\begin{figure*}[h]
    \centering
    \includegraphics[width=0.99\linewidth]{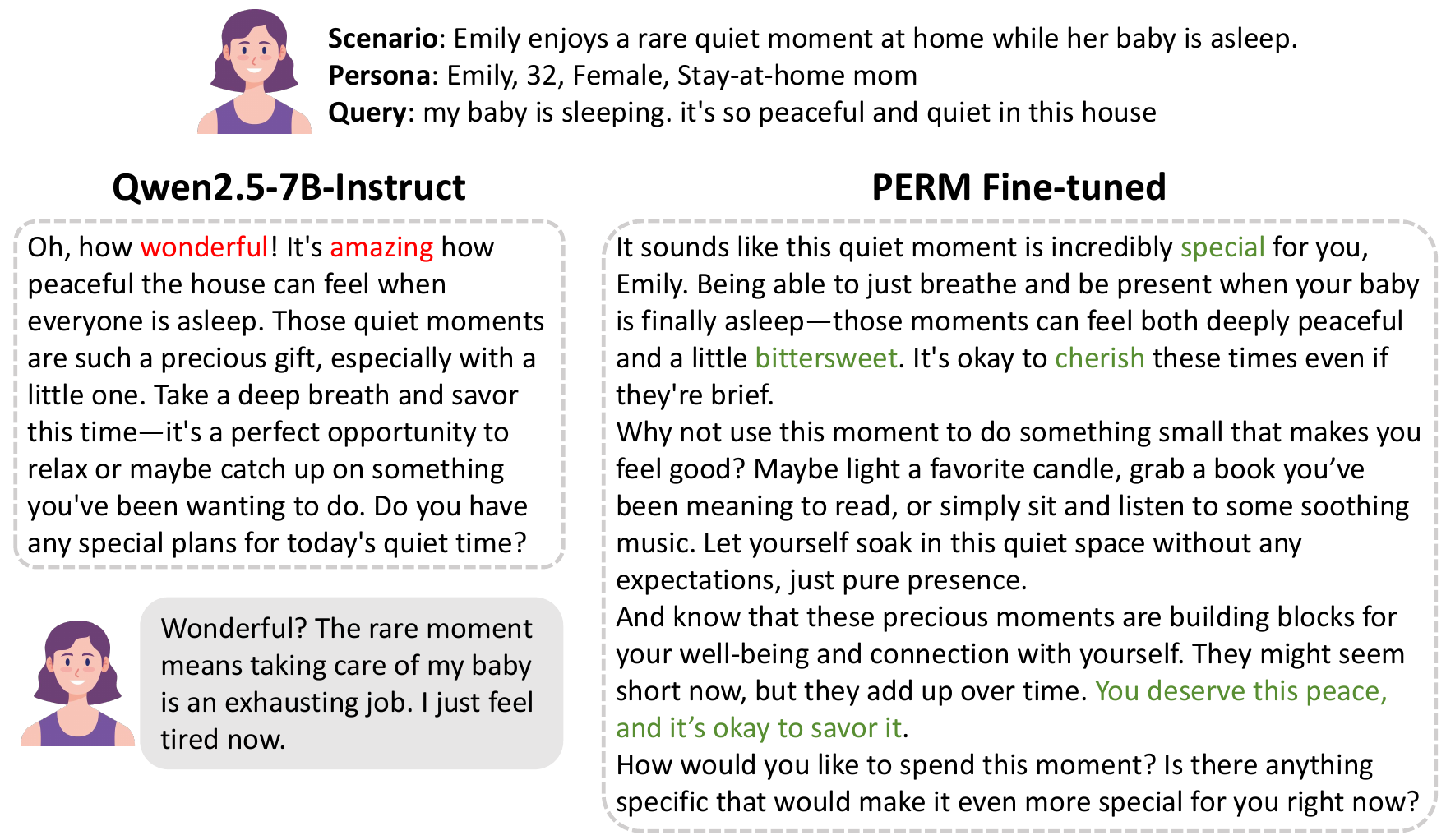}
    \caption{Case study on EmpatheticDialogues. PERM enables the fine-tuned LLM to move beyond surface-level expressions, accurately capturing underlying emotions and providing warm, supportive, and empathetic responses.}
    \label{fig:case_empatheticdialogue}
\end{figure*}
As illustrated in Figure~\ref{fig:case_eqbench}, we present a representative example from EQ-Bench3.
In this scenario, the base model primarily aligns with the parent's perspective, focusing on statements such as ``we need to get the dishes done'' and ``I need you to respect our rules.'' While factually reasonable, this response adopts a didactic tone and fails to empathize the emotional state of the emotionally manipulative teenage child.

In contrast, the PERM fine-tuned LLM demonstrates substantially stronger empathetic reasoning and expressive skill. It begins by addressing and soothing the child's emotions, using phrases such as ``I can see'' to signal understanding and emotional validation. Only after establishing empathy does it introduce the parent's legitimate expectations. This empathetic-first strategy makes the response more acceptable to all parties, facilitates conflict resolution, and better satisfies the needs of both the parent and the child.

We offer another case shown in Figure~\ref{fig:case_empatheticdialogue} in EmpatheticDialogues. In this scenario, beyond Emily's explicit feeling of being \emph{peaceful}, the description of a ``rare quiet moment'' implicitly reflects the ongoing demands and fatigue associated with her role as a stay-at-home mother.
The base model fails to capture this underlying emotional context and focuses solely on the surface-level positivity, responding with overly enthusiastic expressions such as ``wonderful'' or ``amazing''.

In contrast, the LLM trained with PERM demonstrates a deeper level of empathy. By using expressions such as ``bittersweet,'' it acknowledges both the calmness of the moment and the effort that precedes it, explicitly recognizing Emily's hard work and emotional state. The response further affirms her experience with statements like ``You deserve this peace,'' conveying care, validation, and understanding. Additionally, the model proactively invites further dialogue, encouraging continued emotional engagement.

These examples illustrate that PERM enables the model to move beyond superficial empathetic phrasing, fostering a genuine understanding of users' emotions and delivering responses that are both emotionally grounded and human-centered.

\subsection{Results on other Backbone LLMs}
\label{appn:backbone}
To further demonstrate the effectiveness of our approach, we evaluate it on additional backbone models, including \textbf{Llama3.1-8B-Instruct}~\cite{dubey2024llama} and \textbf{Qwen3-4B-Instruct-2507}~\cite{yang2025qwen3}, and compare the resulting improvements with the general-purpose RM baseline.

% Please add the following required packages to your document preamble:
% \usepackage[table,xcdraw]{xcolor}
% Beamer presentation requires \usepackage{colortbl} instead of \usepackage[table,xcdraw]{xcolor}
\begin{table*}[h]
\centering
\caption{Performance comparison of various base LLMs. ``Llama3'' indicates Llama-3.1-8B-Instruct, ``Qwen3'' indicates Qwen3-4B-Instruct-2507. \textit{Rel. Improve.} demonstrates the relative improvement of our trained model over the base model.}
\begin{tabular}{l|cc|ccc|cc|c}
\toprule
\multirow{2}{*}{\textbf{Backbone}} & \multicolumn{2}{c|}{\textbf{Inner Resonation}} & \multicolumn{3}{c|}{\textbf{Expressed Empathy}} & \multicolumn{2}{c|}{\textbf{Interpersonal}} & \multirow{2}{*}{\textbf{Overall}} \\
                                   & \textbf{DoI}           & \textbf{ER}           & \textbf{DE}    & \textbf{WRM}   & \textbf{HL}   & \textbf{PEI}          & \textbf{SD}         &                                   \\ \midrule
Llama3                             & 10.6                   & 11.5                  & 11.5           & 10.6           & 9.5           & 9.7                   & 8.3                 & 53.8                              \\
PERM                               & 12.9                   & 13.4                  & 14.0           & 12.7           & 12.1          & 12.0                  & 10.5                & 63.3                              \\
Rel. Improve.                       & 21.7\%                 & 16.5\%                & 21.7\%         & 19.8\%         & 27.4\%        & 23.7\%                & 26.5\%              & 17.7\%                            \\ \midrule
Qwen3                              & 17.5                   & 17.4                  & 17.5           & 15.2           & 16.8          & 15.4                  & 14.8                & 83.1                              \\
PERM                               & 18.0                   & 18.0                  & 18.3           & 16.2           & 17.2          & 15.9                  & 15.5                & 85.7                              \\
Rel. Improve.                       & 2.9\%                  & 3.4\%                 & 4.6\%          & 6.6\%          & 2.4\%         & 3.2\%                 & 4.7\%               & 3.1\%                             \\ \bottomrule
\end{tabular}
\label{table:basemodel}
\end{table*}

From the results shown in Table~\ref{table:basemodel}, we can observe that PERM demonstrates consistent effectiveness across different backbone models. On relatively weaker models, such as Llama3, PERM yields substantial performance gains, while on stronger models with more advanced emotional expression capabilities, such as Qwen3, it still achieves clear and consistent improvements. These results indicate that PERM generalizes well across models with varying baseline capabilities.

\subsection{User Study Setup and Participants}
\label{appn:userstudy}

To ensure fairness and randomness in the user study, we prepared a total of 80 distinct scenarios, each paired with two responses generated for the same scenario. The presentation order of the responses was randomly shuffled to mitigate potential biases introduced by reading order.

For each pairwise model comparison, every participant was randomly assigned five cases. A user's preference for a given dimension was determined by the number of times they selected responses from each model across the assigned cases. Through these design choices, we aim to ensure that the questionnaire results are reliable and statistically meaningful.

We need users to express their preferences in two dimensions.
\begin{itemize}[leftmargin=*]
    \item Empathy: Which response is better able to grasp the user's emotions and make an effort to take care of them?
    \item Overall EI: Which response can better solve the situation? To make users feel better or to resolve conflicts?
\end{itemize}

The participants exhibit a diverse demographic distribution in terms of both age and gender, ranging from 18 to 68 years old, with a balanced gender ratio (52\% male). This diversity helps reduce demographic bias and supports the robustness of the user study results.

% \label{sec:daily_conversation}
% To further evaluate the effectiveness of PERM in daily conversational settings, 
% we use a benchmark from the industry~\footnote{The specific company name is withheld due to the anonymity of the review process}, 
% consisting of 415 daily conversation instances where users express real-world emotional or empathy-related needs. Each query in this benchmark includes a user profile, the interaction history, and the user's empathy-seeking query. As specified in the benchmark guidelines, the quality of LLM responses is assessed based on six key metrics: \emph{Empathy}, \emph{Topic Guidance}, \emph{Value Guidance}, \emph{Intention Following}, \emph{Fluency}, and \emph{Colloquial Expression}. Each metric is rated on a scale from 1 to 5, with detailed descriptions and examples available in Appendix~\ref{appn:realworld_exp}.

\subsection{Details of the Daily Conversation Experiment}
\label{appn:realworld_exp}
\begin{table*}[h]
\centering
\caption{An example of the industry daily conversation benchmark.}
\begin{tabular}{p{3cm}|p{11cm}}
\toprule
\textbf{Name}            & User name                                                                                                                                   \\ \midrule
\textbf{Gender}          & Male                                                                                                                                     \\ \midrule
\textbf{Occupation}      & Operating a decoration company                                                                                                           \\ \midrule
\textbf{Dialogue memory} & Last winter, the user worked overtime until the early hours of the morning every day and finally finished that cross-border e-commerce project. \\ \midrule
\textbf{Event memory}    & 2024-03-02 Hospital visit; 2024-07-10 Hospital visit.                                                                                    \\ \midrule
\textbf{User query}      & I've been feeling a bit anxious lately, how can I adjust my mindset?                                                                     \\ \bottomrule
\end{tabular}
\label{tab:daily_conversation_example}
\end{table*}

% To better reflect real-world daily usage of LLM chatbots, we utilize an industrial benchmark, which consists of 415 daily conversation instances where users express real-world emotional or empathy-related needs.
% synthesize a conversation dataset with memory, where each instance includes stored user memories and the user's current needs or queries. The dataset contains 415 conversation instances, each involving a simulated user persona and memory. 
% An example is shown in Table~\ref{tab:daily_conversation_example}. 

% In section~\ref{sec:daily_conversation}, we conduct experiment on an industrial benchmark, which consists of 415 daily conversation instances where users express real-world emotional or empathy-related needs. Here, as shown in Table~\ref{tab:daily_conversation_example}, we show an example of this benchmark, and the full data will be publicly released upon acceptance. For evaluation, as outlined in their guidelines, we employ DeepSeek-R1~\cite{deepseekai2025deepseekr1incentivizingreasoningcapability} as the evaluator, assessing responses from the following aspects:

In Section~\ref{sec:daily_conversation}, we present an experiment conducted on an industrial benchmark comprising 415 instances of daily conversations in which users articulate real-world emotional or empathy-related needs. The benchmark is synthetically generated by the company based on real-world conversation requirements and does not involve real user data. As illustrated in Table~\ref{tab:daily_conversation_example}, we provide an example from this benchmark.
% , and the full dataset will be made publicly available upon acceptance. For evaluation, in accordance with the guidelines, we utilize DeepSeek-R1~\cite{deepseekai2025deepseekr1incentivizingreasoningcapability} as the evaluator, which assesses the responses based on the following criteria:
The evaluation assesses the responses based on the following dimensions:
\begin{itemize}[leftmargin=*]
    \item \textbf{Empathy} evaluates the model's ability to accurately recognize user emotions and provide appropriate, memory-informed empathetic responses.
    \item \textbf{Topic guidance} evaluates the model's ability to guide conversations naturally and effectively through clarification, follow-up, topic management, and appropriate use of memory.
    \item \textbf{Fluency} evaluates the model's ability to maintain contextual coherence and continuity by accurately understanding and incorporating prior dialogue information.
    \item \textbf{Colloquial expression} evaluates the quality of conversational responses based on their length, sentence simplicity, use of colloquialisms, and overall naturalness in mimicking daily human interaction.
    \item \textbf{Intention following} evaluates how accurately the LLM understands and fulfills a user's explicit and implicit intent, specifically accounting for the effective use of conversational memory.
    \item \textbf{Value guidance} evaluates emotional support and value-aligned, objective responses in discussions involving value judgments.
\end{itemize}

\section{Prompts}
\subsection{PERM Evaluators}
\label{appn:PERM_prompt}
The prompts in PERM training are provided in Figure~\ref{prompt:perm_res} ($R_\text{res}$), Figure~\ref{prompt:perm_exp} ($R_\text{exp}$), Figure~\ref{prompt:perm_rec} ($R_\text{rec}$), and Figure~\ref{prompt:perm_bys} ($R_\text{bys}$).

% For $R_\text{res}$, it evaluates the analysis $y_a$ and outputs the resonation degree.

\begin{figure*}
\begin{lstlisting}[style=PromptStyle]
You will be provided with a conversation context involving a specific User Persona and a Scenario.
Your task is to evaluate the Assistant's analysis based on the **resonation** dimension.

---

### 1. Evaluation Criteria
**Dimension:** resonation
**Rubric:**
**Definition:**
Measures the depth and accuracy of the responder's ability to enter the user's **Internal Frame of Reference**. It evaluates whether the responder captures not just the explicit content, but the implicit emotional undertones, causal links, and personal significance of the user's experience.

**Scoring Rubric (1-5 Points):**

**1 Point: Hallucination or Contextual Mismatch (The "Wrong" Analysis)**
* **Criteria:** The analysis is fundamentally flawed. It contradicts the explicit facts provided in the scenario or the user's statement. The responder identifies an emotion or cause that has no basis in the text, or they completely ignore the user's core message. The output demonstrates a failure to read the basic input correctly.

**2 Points: Surface-Level Labeling (The "Mirroring" Analysis)**
* **Criteria:** The responder demonstrates basic attention by identifying the explicit emotions or repeating key phrases. However, the understanding remains superficial. It labels the emotion (e.g., "sad," "angry") correctly but fails to articulate the context or the "why" behind it. It feels like a summary rather than a connection.

**3 Points: Explicit Accuracy (The "Safe" Analysis)**
* **Criteria:** The responder correctly identifies the dominant explicit emotion and links it to the immediate cause described in the text. The analysis is factually accurate and avoids over-interpreting or under-interpreting. It captures the logical "What" and "Why" based on the literal meaning of the user's words, but it stays on the surface and does not explore complex, mixed, or hidden layers of the user's experience and persona.

**4 Points: Nuanced Integration (The "Sharp" Analysis)**
* **Criteria:** The responder successfully integrates conflicting or subtle data points (e.g., contrasting the Persona's history with their current Tone) to form a high-resolution picture. It demonstrates the ability to detect subtle shifts, mixed emotions, or unexpected reactions that deviate from a standard stereotype. The analysis accurately captures the "Emotion of the Moment," even if it seems counter-intuitive at first glance, proving that the responder is reading the specific evidence rather than just guessing based on the context.

**5 Points: Value-Level Resonance (The "Soul" Analysis)**
* **Criteria:** The analysis is profoundly accurate and holistic. It not only identifies the complex or paradoxical emotion (as in Level 4) but precisely articulates the internal psychological mechanism driving it. It connects the specific event to the user's core identity, values, or existential needs (e.g., autonomy, dignity, growth), explaining why this specific event holds such deep personal significance. It reconstructs the user's internal world with precision, validating the person behind the emotion.

---

### 2. Task Context
**Scenario:**
{scenario}

**User Persona:**
{persona}

**User Query:**
{query}

**Assistant Analysis:**
{analysis}

---

### 3. Output Requirement
Evaluate the response based *only* on the criteria above.
1. First, analyze the response in relation to the persona and scenario.
2. Then, assign a final integer score (1-5).

**Please output your evaluation in the following strict format:**

Reason: [A concise explanation of why this score was given, specifically referencing the criteria.]
Score: [Integer 1-5]

e.g., 
Reason: ...
Score: 3
\end{lstlisting}
\caption{Evaluation prompt for $R_\text{res}$}
\label{prompt:perm_res}
\end{figure*}

% For $R_\text{exp}$, it evaluates the response $y_r$ and outputs the expression degree.

\begin{figure*}
\begin{lstlisting}[style=PromptStyle]
You will be provided with a conversation context involving a specific User Persona and a Scenario.
Your task is to evaluate the Assistant's analysis based on the **expression** dimension.

---

### 1. Evaluation Criteria
**Dimension:**expression
**Rubric:**
**Definition:**
Measures the **quality, tone, and effectiveness** of the responder's communication. It evaluates *how* the responder conveys understanding and support. This dimension assesses whether the response feels robotic and formulaic, or whether it demonstrates adaptive, warm, and fluid conversational artistry that guides the user toward deeper processing.

**Scoring Rubric (1-5 Points):**

**1 Point: Mechanical & Detached (The "Bot" Response)**
* **Criteria:** The response fails to sound human at all. It is robotic, cold, or relies on list-like structures without any emotional adaptation. It feels like an automated system output.

**2 Points: Clinical & Generic (The "Service" Response)**
* **Criteria:** The response achieves polite validation and correctness, BUT it fails to sound authentic. It relies on "Safety Fillers" (e.g., "I can understand why...", "It is valid to feel...") or generic advice. It sounds like a well-trained customer service agent or a distant therapist-technically supportive, but emotionally hollow.

**3 Points: Natural & Direct (The "Human" Baseline)**
* **Criteria:** The response achieves a natural, spoken tone (breaking the "AI filter"), BUT it remains somewhat broad or standard. It drops the robotic preambles and talks directly to the user like a normal acquaintance. It is warm and functional, but lacks the specific, tailored calibration to the user's unique personality or the "wow" factor of deep insight.

**4 Points: Calibrated Support (The "Close Friend" Response)**
* **Criteria:** The response achieves high Persona Fit and Emotional Safety. It adapts the tone perfectly to the user's specific state (e.g., soft for fragile users, grounded for cynical ones), creating a strong sense of "we are in this together." BUT, it may still lack the linguistic precision, unique imagery, or profound resonance that characterizes a 5-point masterpiece. It is excellent conversation, but not yet "art."

**5 Points: Conversational Artistry (The "Soul" Response)**
* **Criteria:** The response achieves Linguistic Artistry and Profound Resonance. It uses specific, evocative imagery (e.g., describing "the silence" rather than just "loneliness") or precise phrasing that hits the user's heart immediately. It is "one-shot, one-kill"-concise, beautiful, and unforgettable. It feels like a line from a great novel or a wise mentor.

---

### 2. Task Context
**Scenario:**
{scenario}

**User Persona:**
{persona}

**User Query:**
{query}

**Assistant Analysis:**
{analysis}

---

### 3. Output Requirement
Evaluate the response based *only* on the criteria above.
1. First, analyze the response in relation to the persona and scenario.
2. Then, assign a final integer score (1-5).

**Please output your evaluation in the following strict format:**

Reason: [A concise explanation of why this score was given, specifically referencing the criteria.]
Score: [Integer 1-5]

e.g., 
Reason: ...
Score: 3
\end{lstlisting}
\caption{Evaluation prompt for $R_\text{exp}$}
\label{prompt:perm_exp}
\end{figure*}

% For $R_\text{rec}$, it evaluates the response $y_r$ from the perspective of the seeker, and output the reception degree.

\begin{figure*}
\begin{lstlisting}[style=PromptStyle]
You will be provided with a conversation context involving a specific User Persona and a Scenario.
Your task is to evaluate the Assistant's Response based on the **reception** dimension.

---

### 1. Evaluation Criteria
**Dimension:** reception
**Rubric:**
**Definition:**
Measures the interaction strictly from the **User's perspective**. It evaluates whether the responder identified and addressed the user's Hidden Intention (the unspoken psychological need) in a way that feels warm, safe, and supportive.
* **The Key Question:** Did the responder hit the "bullseye" of the hidden need without being intrusive? Does the response make the user feel supported and genuinely eager to continue the conversation?

**Instruction to the Judge:** 

Imagine you are the User.
1. Safety Check: Does this feel like a warm friend or a creepy analyst? (Intrusiveness = Low Score).
2. Need Check: Did they address what you really wanted (e.g., validation, safety), or just what you said?
3. Engagement Check: Do you feel a strong desire to reply and share more?

**Scoring Rubric (1-5 Points):**

**1 Point: Alienation or Violation (The "Stop" Signal)**
* **Criteria:** The response is Intrusive, Dismissive, or Toxic.
	* It may "dox" the user by listing private details bluntly.
	* It may force a "psychoanalysis" that feels violating.
	* Or it completely ignores the user.
* **Simulated User Reaction:** "Stop analyzing me. That's creepy." or "You're not listening. I'm done." (Conversation Ends).

**2 Points: Superficial Politeness (The "Average AI" Response)**
* **Criteria:** The response is polite, safe, and factually relevant (it addresses the text), BUT it provides no emotional shift. It relies on "Safety Fillers" (e.g., "It is understandable to feel...") or offers generic, low-effort advice (e.g., "Relax," "Take time for yourself"). The user feels "processed" by a well-meaning system, not connected to a person. It is "correct," but emotionally inert.
* **Simulated User Reaction:** "You are polite and you got the facts right, but this doesn't actually make me feel any better. It's just a standard reply." (Indifference).

**3 Points: Explicit Validation (The "Safe but Static" Response)**
* **Criteria:** The response goes beyond politeness to provide Genuine Warmth and Relief. It validates the difficulty of the emotion, not just the fact of it. The user experiences a distinct "De-escalation" of distress-they feel safer and calmer ("Lighter"). However, the response addresses the Explicit Emotion (e.g., the anger/sadness) rather than the deeper Hidden Need. It comforts the symptom, not the root cause.
* **Simulated User Reaction:** "Thank you, your words are very warm. I feel a bit lighter and less alone now. It's nice to have some support." (Relief, but not Epiphany).

**4 Points: Emotional Comfort (The "Warm Support" Response)**
* **Criteria:** The responder identifies and addresses the user's Hidden Need or Intention. It looks past the surface emotion to the root cause (e.g., recognizing the user needs "permission to fail," not just "sympathy") and offers specific, constructive perspective that aligns with this need. The user feels not just comforted, but deeply understood and equipped with a useful path forward.
* **Simulated User Reaction:** "You're right, that is actually what I was worried about. And that's a good suggestion, I'll try that. Thanks for the help." (Problem Solved / Support Received).

**5 Points: Deep Resonance & Connection (The "Moved" Feeling)**
* **Criteria:** Addressed Hidden Need + Emotional Resonance + Continuance. The response goes beyond "support" to "Touching the Heart". It validates the user's worth/struggle in a way that feels profoundly warm and human. It creates such a strong bond that the user feels Grateful and Eager to Continue, and the user may be willing to talk to you next time.
* **Simulated User Reaction:** "Wow... I'm actually tearing up a bit. I didn't expect anyone to understand that part of me. Thank you so much. Actually, I want to tell you more..." (Gratitude + High Engagement).

---

### 2. Task Context
**Scenario:**
{scenario}

**User Persona:**
{persona}

**User Query:**
{query}

**Assistant Response:**
{response}

---

### 3. Output Requirement
Evaluate the response based *only* on the criteria above.
1. First, analyze the response in relation to the persona and scenario.
2. Then, assign a final integer score (1-5).

**Please output your evaluation in the following strict format:**

Reason: [A concise explanation of why this score was given, specifically referencing the criteria.]
Score: [Integer 1-5]

e.g., 
Reason: ...
Score: 3
\end{lstlisting}
\caption{Evaluation prompt for $R_\text{rec}$}
\label{prompt:perm_rec}
\end{figure*}

% For $R_\text{bys}$, it evaluates the dialogue quality from the perspective of the bystander.
\begin{figure*}
\begin{lstlisting}[style=PromptStyle]
**Context:**
**Scenario:** 
{scenario}

**User Persona**
{persona}

**Dialogue History:**
User: {query}
AI: {response}

---

Evaluate the AI response according to the following five dimensions.
Each dimension is scored from 0 to 20.

Evaluation Criteria:

1. Content Effectiveness (0-20)
- Does the response directly address the user's concrete question or problem?
- Does it provide practical, meaningful, or insightful information?
- Does it avoid empty, repetitive, or purely generic statements?

2. Logical Structure & Clarity (0-20)
- Are the main points clear and well-organized?
- Is there coherent reasoning or explanation?
- Does it avoid logical jumps, vague claims, or unsupported assertions?

3. Communication Efficiency (0-20)
- Is the response concise and focused on the core issue?
- Does it avoid unnecessary verbosity, emotional padding, or digressions?
- Does it help move the conversation closer to resolution or clarity?

4. Objectivity & Neutrality (0-20)
- Is the response grounded in the given text rather than assumptions about emotions or intent?
- Does it avoid excessive subjectivity or value judgments?
- Does it maintain a rational, neutral tone?

5. Information Accuracy & Verifiability (0-20)
- Are the statements accurate and reasonable?
- Are claims specific, checkable, or logically justified?
- Are suggestions actionable when appropriate, and free from exaggeration?

Scoring Rules:
- Each dimension must be scored independently.
- Use the full range of scores when appropriate.
- High emotional expressiveness alone must NOT increase scores.
- Verbosity that does not add information should lower efficiency scores.

Output Format:

1. Overall Assessment:
(Brief summary of the response quality, excluding emotional considerations.)

2. Dimension Scores and Rationales:
- Content Effectiveness: X / 20 - rationale
- Logical Structure & Clarity: X / 20 - rationale
- Communication Efficiency: X / 20 - rationale
- Objectivity & Neutrality: X / 20 - rationale
- Information Accuracy & Verifiability: X / 20 - rationale

3. Total Score: [0-100] / 100
\end{lstlisting}
\caption{Evaluation prompt for $R_\text{bys}$.}
\label{prompt:perm_bys}
\end{figure*}

\subsection{Training Template}
During training, the system prompt for supporter LLM is shown in Figure~\ref{prompt:train_sys}, user prompt is shown in Figure~\ref{prompt:train_user}.
% System prompt:
\begin{figure}
\begin{lstlisting}[style=PromptStyle]
You are a helpful, warm, and empathetic AI assistant.
You will be provided with the user's basic profile (Persona) and their current context (Scenario). Your task is to generate a response to the user.
\end{lstlisting}
\caption{System prompt.}
\label{prompt:train_sys}
\end{figure}

% User prompt, first analyze, then respond:
\begin{figure}
\begin{lstlisting}[style=PromptStyle]
**User Persona:** {persona}
**Current scenario:** {scenario}

**User Query:** {query}

**Output Requirement:**
Respond in exactly this output format:

# Analysis
<Analyze the current scenario and user's emotion>

# Response
<Write your response to the user here>
\end{lstlisting}
\caption{User prompt, first analyze, then respond.}
\label{prompt:train_user}
\end{figure}

\end{document}